\documentclass{article}

\usepackage[preprint]{neurips_data_2024}

\usepackage[utf8]{inputenc} %
\usepackage[T1]{fontenc}    %
\usepackage{hyperref}    %
\usepackage{url}            %
\usepackage{booktabs}       %
\usepackage{amsfonts}       %
\usepackage{nicefrac}       %
\usepackage{microtype}      %
\usepackage{xcolor}         %
\usepackage{amssymb}
\usepackage{natbib}
\usepackage{wrapfig,lipsum,booktabs}
\usepackage{caption}
\usepackage{etoc}
\usepackage{appendix}
\usepackage[para]{footmisc}

\setcitestyle{square,numbers,comma}

\DeclareCaptionLabelFormat{andtable}{#1~#2  \&  \tablename~\thetable}
\newcommand*\samethanks[1][\value{footnote}]{\footnotemark[#1]}
\hypersetup{
colorlinks=true,
linkcolor=blue,
citecolor=blue,
filecolor=magenta,
urlcolor=blue
}

\title{SMPLOlympics: Sports Environments for Physically Simulated Humanoids}

\usepackage{authblk}
\author{%
Zhengyi Luo$^{1}$ \thanks{\tiny{Equal Contribution}} \quad Jiashun Wang$^{1}$ \samethanks \quad Kangni Liu$^{1}$ \samethanks \quad Haotian Zhang$^2$ \quad Chen Tessler$^2$ \quad Jingbo Wang  \quad Ye Yuan$^2$ \quad Jinkun Cao$^1$ \quad Zihui Lin$^1$ \quad Fengyi Wang$^1$ \quad
Jessica Hodgins$^1$ \quad Kris Kitani$^1$ \\
        $^1$Carnegie Mellon University; $^2$Nvidia\\
        {\tt\small \href{https://smplolympics.github.io/SMPLOlympics}{https://smplolympics.github.io/SMPLOlympics}} \\
}
\usepackage{amsfonts}
\usepackage{bm}
\usepackage{amsmath}
\usepackage{mathtools}
\usepackage{multirow}
\usepackage{booktabs}
\usepackage{caption}
\usepackage{enumitem}
\usepackage{wrapfig,lipsum,booktabs}
\usepackage{caption}
\usepackage{bbm}
\usepackage{multirow}
\usepackage{pifont}
\usepackage[linesnumbered,ruled,vlined]{algorithm2e}

\SetCommentSty{mycommfont}
\SetAlCapFnt{\small} 
\usepackage{etoc}
\SetAlgorithmName{Algo}{Algo}{}

\renewcommand{\paragraph}[1]{{\vspace{1mm}\noindent \bf #1}.}

\newcommand{\reals}{\mathbb{R}}

\newcommand{\decoder}{\bs{\mathcal{D}}_{\text{PULSE}}}
\newcommand{\prior}{\bs{\mathcal{P}}_{\text{PULSE}}}

\newcommand{\link}{https://smplolympics.github.io/SMPLOlympics}
\newcommand{\name}{\text{SMPLOlympics}}

\newcommand{\policytask}{{\pi_{\text{task}}}}

\newcommand{\rewardfunc}{\bs{\mathcal{R}}}

\newcommand{\rewardfuncfencing}{\rewardfunc^{\text{fencing}}}

\newcommand{\rewardracket}{r^{\text{racket}}_t}
\newcommand{\rewardball}{r^{\text{ball}}_t}
\newcommand{\rewardfacing}{r^{\text{facing}}_t}
\newcommand{\rewardvel}{r^{\text{vel}}_t}
\newcommand{\rewardstrike}{r^{\text{strike}}_t}
\newcommand{\rewardpoint}{r^{\text{point}}_t}

\newcommand{\ballp}{{\bs{p}^{\text{ball}}_{t}}}

\newcommand{\ballpxy}{{\bs{p}^{\text{ball}}_{t,xy}}}
\newcommand{\ballv}{{\bs{v}^{\text{ball}}_{t}}}
\newcommand{\closertogoalxy}{{g^{\text{ball-to-goal}}_{t}}}
\newcommand{\racketp}{{\bs{p}^{\text{racket}}_{t}}}
\newcommand{\clubp}{{\bs{p}^{\text{club}}_{t}}}
\newcommand{\targetp}{{\bs{p}^{\text{tar}}_{t}}}
\newcommand{\targetpxy}{{\bs{p}^{\text{tar}}_{t,xy}}}
\newcommand{\ballpprev}{{\bs{p}^{\text{ball}}_{t - 1}}}

\newcommand{\opptroot}{{\bs{p}^{\text{opp-root}}_t}}
\newcommand{\oppt}{{\bs{p}^{\text{opp}}_t}}

\newcommand{\oppv}{{\bs{v}^{\text{opp}}_t}}
\newcommand{\oppttarget}{{\bs{p}^{\text{opp-target}}_t}}
\newcommand{\simtsword}{{\bs{p}^{\text{sword}}_{t}}}
\newcommand{\contact}{{\bs{{c}}_{t}}}
\newcommand{\contactopp}{{\bs{{c}}^{\text{opp}}_{t}}}
\newcommand{\bbox}{\bs{b}}

\newcommand{\simt}{{\bs{{p}}_{t}}}
\newcommand{\simr}{{\bs{\theta}_{t}}}
\newcommand{\simthand}{{\bs{{p}}_{t}^{\text{hand}}}}

\newcommand{\goalstate}{{\bs{s}^{\text{g}}_t}}

\newcommand{\goalstatesoccer}{{\bs{s}^{\text{g-soccer}}_t}}
\newcommand{\goalstatekick}{{\bs{s}^{\text{g-kick}}_t}}

\newcommand{\goalstatefencing}{{\bs{s}^{\text{g-fencing}}_t}}
\newcommand{\goalstateboxing}{{\bs{s}^{\text{g-boxing}}_t}}
\newcommand{\goalstategolf}{{\bs{s}^{\text{g-golf}}_t}}
\newcommand{\goalstatehurdling}{{\bs{s}^{\text{g-hurdling}}_t}}
\newcommand{\goalstatehighjump}{{\bs{s}^{\text{g-high\_jump}}_t}}
\newcommand{\goalstatejavelin}{{\bs{s}^{\text{g-javelin}}_t}}
\newcommand{\goalstatelongjump}{{\bs{s}^{\text{g-long\_jump}}_t}}
\newcommand{\goalstatetabletennis}{{\bs{s}^{\text{g-table\_tennis}}_t}}
\newcommand{\goalstatetennis}{{\bs{s}^{\text{g-tennis}}_t}}

\newcommand{\selfstate}{{\bs{s}^{\text{p}}_t}}

\newcommand{\state}{{\bs{s}_t}}
\newcommand{\action}{{\bs{a}_t}}

\newcommand{\objp}{\bs{q}^{\text{obj}}_{t}}

\newcommand{\objr}{\bs{\theta}^{\text{obj}}_{t}}
\newcommand{\objt}{\bs{p}^{\text{obj}}_{t}}
\newcommand{\objv}{\bs{v}^{\text{obj}}_{t}}
\newcommand{\objav}{\bs{\omega}^{\text{obj}}_{t}}

\newcommand{\simp}{{\bs{{q}}_{t}}}
\newcommand{\simv}{{\bs{\dot{q}}_{t}}}
\newcommand{\simvs}{{\bs{\dot{q}}_{1:T}}}
\newcommand{\simav}{{\bs{{\omega}}_{t}}}
\newcommand{\simlv}{{\bs{v}_{t}}}

\newcommand{\bs}[1]{\boldsymbol{#1}}

\newcommand{\rewardfuncsoccerpenalty}{\rewardfunc^{\text{soccer-kick}}}
\newcommand{\rewardfuncsoccertwovtwo}{\rewardfunc^{\text{soccer-match}}}
\newcommand{\rewardfuncfreethrow}{\rewardfunc^{\text{free-throw}}}
\newcommand{\simtball}{{\bs{p}^{\text{ball}}_{t}}}
\newcommand{\simballv}{{\bs{\dot{q}}^{\text{ball}}_t}}
\newcommand{\simballlv}{{\bs{{v}}^{\text{ball}}_t}}
\newcommand{\simballlvdesired}{{\bs{{v}}^{\text{ball-desired}}_t}}
\newcommand{\simballlvdesiredxy}{{\bs{{v}}^{\text{ball-desired}}_{t,xy}}}
\newcommand{\simballlvdesiredz}{{\bs{{v}}^{\text{ball-desired}}_{t,z}}}
\newcommand{\simtgoalpost}{{\bs{p}^{\text{goal-post}}_{t}}}
\newcommand{\simtgoaltarget}{{\bs{p}^{\text{goal-target}}_{t}}}

\newcommand{\allytroot}{{\bs{p}^{\text{ally-root}}_t}}
\newcommand{\opptbodytarget}{{\bs{p}^{\text{opp-target}}_{t}}}

\newcommand{\rootpos}{{\bs{p}^{\text{root}}_{t}}}
\newcommand{\rootposprev}{{\bs{p}^{\text{root}}_{t - 1}}}

\newcommand{\javelinpos}{{\bs{p}^{\text{javelin}}_{t}}}
\newcommand{\javelinpose}{{\bs{q}^{\text{javelin}}_{t}}}
\newcommand{\javelinposedefault}{{\bs{q}^{\text{javelin-default}}_{t}}}
\newcommand{\righthandpos}{{\bs{p}^{\text{right-hand}}_{t}}}
\newcommand{\handpos}{{\bs{p}^{\text{hand}}_{t}}}

\newcommand{\rewardfunchighjump}{\bs{\mathcal{R}}^\text{high jump}}

\newcommand{\rewardfunclongjump}{\bs{\mathcal{R}}^\text{long jump}}
\newcommand{\rewardfuncgolf}{\bs{\mathcal{R}}^\text{golf}}

\newcommand{\rewardfunchurdling}{\bs{\mathcal{R}}^\text{hurdling}}
\newcommand{\rewardfunctennis}{\bs{\mathcal{R}}^\text{tennis}}
\newcommand{\rewardfunctabletennis}{\bs{\mathcal{R}}^\text{table tennis}}

\newcommand{\rewardfuncjavelin}{\bs{\mathcal{R}}^\text{javelin}}

\newcommand{\highjumpposgoal}{\bs{p}^{\text{g-high jump}}}
\newcommand{\highjumppos}{{\bs{p}^{\text{p}}_{t}}}
\newcommand{\highjumpposprev}{{\bs{p}^{\text{p}}_{t-1}}}
\newcommand{\highjumpposx}{{\bs{p}^{\text{p}}_{t,x}}}
\newcommand{\highjumpposz}{{\bs{p}^{\text{p}}_{t,z}}}

\newcommand{\longjumpposgoal}{\bs{p}^{\text{g-long jump}}}
\newcommand{\longjumppos}{{\bs{p}^{\text{p}}_{t}}}

\newcommand{\longjumpposprev}{{\bs{p}^{\text{p}}_{t-1}}}
\newcommand{\longjumpposx}{{\bs{p}^{\text{p}}_{t,x}}}
\newcommand{\longjumpposz}{{\bs{p}^{\text{p}}_{t,z}}}
\newcommand{\longjumpvelocityx}{{\bs{v}^{\text{p}}_{t,x}}}

\newcommand{\hurdlingposgoal}{\bs{p}^{\text{g-hurdling}}}
\newcommand{\hurdlingpos}{{\bs{p}^{\text{p}}_{t}}}
\newcommand{\hurdlingposprev}{{\bs{p}^{\text{p}}_{t-1}}}

\DeclareMathOperator*{\argmin}{argmin}
\begin{document}

\maketitle

\begin{abstract}
\vspace{-4mm}
We present $\name$, a collection of physically simulated environments that allow humanoids to compete in a variety of Olympic sports. Sports simulation offers a rich and standardized testing ground for evaluating and improving the capabilities of learning algorithms due to the diversity and physically demanding nature of athletic activities. As humans have been competing in these sports for many years, there is also a plethora of existing knowledge on the preferred strategy to achieve better performance. To leverage these existing human demonstrations from videos and motion capture, we design our humanoid to be compatible with the widely-used SMPL and SMPL-X human models from the vision and graphics community. We provide a suite of individual sports environments, including golf, javelin throw, high jump, long jump, and hurdling, as well as competitive sports, including both 1v1 and 2v2 games such as table tennis, tennis, fencing, boxing, soccer, and basketball. Our analysis shows that combining strong motion priors with simple rewards can result in human-like behavior in various sports. By providing a unified sports benchmark and baseline implementation of state and reward designs, we hope that $\name$ can help the control and animation communities achieve human-like and performant behaviors.

\end{abstract}

\etocdepthtag.toc{mtchapter}
\etocsettagdepth{mtchapter}{subsection}
\etocsettagdepth{mtappendix}{none}

\section{Introduction}

Competitive sports, much like their role in human society, offer a standardized way of measuring the performance of learning algorithms and creating emergent human behavior. While there exist isolated efforts to bring individual sport into physics simulation \cite{Liu2021-iz, Zhang2023-ox, liu2018learning, yin2021discovering, Won2021-sn, wang2024physpingpong}, each work uses a different humanoid, simulator, and learning algorithm, which prevents unified evaluation. Their specially built humanoids also make it difficult to acquire compatible motion data, as retargeting might be required to translate motion to each humanoid. Building a collection of simulated sports environments that uses a shared humanoid embodiment and training pipeline is challenging, as it requires expert knowledge in humanoid design, reinforcement learning (RL), and physics simulation. 

These challenges have led to previous benchmarks and simulated environments \cite{1606.01540, tunyasuvunakool2020dm_control} focusing mainly on locomotion tasks for humanoids. While these tasks (e.g., moving forward, getting up from the ground, traversing terrains) are as benchmarks, they lack the depth and diversity needed to induce a wide range of behaviors and strategies. As a result, these environments do not fully exploit the potential of humanoids to discover actions and skills found in real-world human activities.

Another important aspect of working with simulated humanoids is the ease of obtaining human demonstrations. The resemblance to the human body makes humanoids capable of performing a diverse set of skills; a human can also easily judge the strategies used by humanoids. Curated human motion can be used either as motion prior \cite{Peng2018-fu, Peng2021-xu, Tessler_undated-zi} or in evaluation protocols. Thus, having an easy way to obtain new human motion data compatible with the humanoid, either from motion capture (MoCap) or videos, is critical for simulated humanoid environments.

In this work, we propose $\name$, a collection of physically simulated environments for a variety of Olympic sports. $\name$ offers a wide range of sports scenarios that require not only locomotion skills, but also manipulation, coordination, and planning.  Unified under one humanoid embodiment, our environments provide a rich set of challenges for developing and testing embodied agents. We use humanoids compatible with the SMPL family of models, which enables the direct conversion of human motion in the SMPL format to our humanoid.  For tasks that require articulated fingers, we use SMPL-X~\cite{SMPL-X:2019} based humanoid which has a much higher degree of freedom (DOF); for tasks that do not need hands, we use SMPL~\cite{Bogo2016-kn}. As popular human models, the SMPL family of models is widely adopted in the vision and graphics community, which provides us with access to human pose estimation methods \cite{ye2023decoupling} capable of extracting coherent motion from videos. The existing large-scale human motion dataset \cite{Mahmood2019-ki} in the SMPL format also helps build general-purpose motion representation for humanoids \cite{Luo2023-er}. 

Our sports environments support both individual and competitive sports, providing a comprehensive platform for testing and benchmarking. For individual sports, we include activities such as golf, javelin throw, high jump, long jump, and hurdling. Competitive sports in our suite include 1v1 games such as ping pong, tennis, fencing, and boxing, as well as team sports such as soccer and basketball. To facilitate benchmarking, we also include tasks such as penalty kicks (for soccer) and ball-target hitting (for ping-pong and tennis) that are easy to measure performance. To demonstrate the importance of human demonstrations, we extract motion from videos using off-the-shelf pose estimation methods, and show that using human motion data as motion prior can \cite{Peng2021-xu} significantly improves human likeness in the resulting motion. We also test recent motion representations in simulated humanoid control using hierarchical RL \cite{Luo2023-er}, and show that a learned motion representation combined with simple rewards can lead to many versatile human-like behaviors to achieve impressive sports results (\ie discovering the Fosbury way for high jump). 

In conclusion, our contributions are: (1) we propose $\name$, a collection of simulated environments that allow humanoids to compete in 10 Olympic sports; (2) we provide a pipeline to extract human demonstration data from videos and show their effectiveness in helping build human-like strategies in simulated sports; (3) we provide the starting state and reward designs for each sport, benchmark state-of-the-art algorithms, and show that simple rewards combined with a strong motion prior can lead to impressive sports feats.

\section{Related Works}

\paragraph{Simulated Humanoid Sports}
Simulated humanoid sports can help generate animations and explore optimal sports strategies. Research has focused on various individual sports within simulated environments, including tennis~\cite{Zhang2023-ox}, table tennis~\cite{wang2024physpingpong}, boxing \cite{Won2021-sn, Zhu2023-nn}, fencing \cite{Won2021-sn}, basketball dribbling \cite{liu2018learning, wang2023physhoi} and soccer \cite{xie2022learning, Liu2021-iz}. These studies leverage human motion to achieve human-like behaviors, using it to acquire motor skills \cite{Liu2021-iz, Won2021-sn} or establish motion prior \cite{Zhang2023-ox}. However, the diversity in humanoid definitions across studies makes it difficult to aggregate additional human demonstration data due to the need for retargetting. Furthermore, the task-specific training pipelines in these studies are hard to generalize to new sports. In contrast, $\name$ provides a unified benchmark employing a consistent humanoid model and training pipeline across all sports. This standardization not only facilitates extension to more sports, but also simplifies benchmarking learning algorithms.

\paragraph{Simulated RL Benchmarks}
Simulated full-body humanoids provide a valuable platform for studying embodied intelligence due to their close resemblance to real-world human behavior and physical interactions. Current RL benchmarks \cite{1606.01540, tunyasuvunakool2020dm_control, Makoviychuk2021-sx} often focus on locomotion tasks such as moving forward and traversing terrain. $\texttt{dm\_control}$~\cite{tunyasuvunakool2020dm_control}  and OpenAI~\cite{1606.01540} Gym only include locomotion tasks. ASE~\cite{Peng2022-vr} includes results for five tasks based on mocap data, which involve mainly simple locomotion and sword-swinging actions. These tasks lack the complexity required to fully exploit the capabilities of simulated humanoids. Sports scenarios require agile motion and strategic teamwork. They are also easily interpretable and provide measurable outcomes for success. A concurrent work, HumanoidBench \cite{sferrazza2024humanoidbench} employs a commercially available humanoid robot in simulation to address 27 locomotion and manipulation tasks. Unlike HumanoidBench, ours targets competitive sports and uses available human demonstration data to enhance the learning of human-like behaviors. This emphasis is essential, as without human demonstrations, behaviors developed in benchmarks can often appear erratic, nonhuman-like, and inefficient.

\paragraph{Humanoid Motion Representation}
Adversarial learning has proven to be a powerful method for using human reference motions to enhance the naturalness of humanoid animations \cite{Peng2021-xu, xu2023composite, bae2023pmp}. Due to the high DoF in humanoids and the inherent sample inefficiency of RL training, efforts have focused on developing motion primitives \cite{Hasenclever2020-zg, Merel2018-bq, Haarnoja2018-yo, Rao2021-ya} and motion latent spaces \cite{dou2023c, Peng2022-vr, Tessler_undated-zi}. These techniques aim to accelerate training and provide human-like motion priors. Notably, approaches such as ASE \cite{Peng2022-vr}, CASE \cite{dou2023c}, and CALM \cite{Tessler_undated-zi} utilize adversarial learning objectives to encourage mapping between random noise and realistic motor behavior. Furthermore, methods such as ControlVAE \citep{Yao2022-as}, NPMP \citep{Merel2018-bq}, PhysicsVAE \citep{Won2022-jy}, NCP \citep{Zhu2023-nn}, and PULSE \cite{Luo2023-er} leverage the motion imitation task to acquire and reuse motor skills for the learning of downstream tasks. In this work, we study AMP \cite{Peng2021-xu} and PULSE \cite{Luo2023-er} as exemplary methods to provide motion priors. Our findings demonstrate that a robust motion prior, combined with straightforward reward designs, can effectively induce human-like behaviors in solving complex sports tasks.

\section{Problem Formulation}

We define the full-body human pose as $\simp \triangleq (\simr, \simt)$, consisting of 3D joint rotations $\simr \in \reals^{J \times 6}$ and positions $\simt \in \reals^{J \times 3}$ of all $J$ joints on the humanoid, using the 6 DoF rotation representation \citep{Zhou2019-lj}. To define velocities $\simvs$, we have $\simv \triangleq (\simav, \simlv)$ as angular $\simav \in \reals^{J \times 3}$ and linear velocities $\simlv \in \reals^{J \times 3}$. If an object is involved (\eg javelin, football, ping-pong ball), we define their 3D trajectories $\objp$ using object position $\objt$, orientation $\objr$, linear velocity $\objv$, and angular velocity $\objav$. As a notation convention, we use $\widehat{\cdot}$ to denote the ground truth kinematic quantities from Motion Capture (MoCap) and normal symbols without accents for values from the physics simulation. 

\paragraph{Goal-conditioned Reinforcement Learning for Humanoid Control}
We define each sport using the general framework of goal-conditioned RL. Namely, a goal-conditioned policy $\policytask$ is trained to control a simulated humanoid competing in a sports environment. The learning task is formulated as a Markov Decision Process (MDP) defined by the tuple ${\mathcal M}=\langle \mathcal{\bs S}, \mathcal{ \bs A}, \mathcal{ \bs T}, \rewardfunc, \gamma\rangle$ of states, actions, transition dynamics, reward function, and discount factor. The simulation determines the state $\state \in \mathcal{ \bs S}$ and transition dynamics $\mathcal{ \bs T}$, where a policy computes the action $\action$.  The state $\state$ contains the proprioception $\selfstate$ and the goal state $\goalstate$. Proprioception is defined as $\selfstate \triangleq (\simp, \simv)$, which contains the 3D body pose $\simp$ and velocity $\simv$. We use $\bbox$ to indicate the boundary of the arena to which a sport is limited. All values are normalized with respect to the humanoid heading (yaw).

\section{SMPLOlympics: sports environments For Simulated Humanoids}
\begin{figure*}
\vspace{-1.5mm}
\begin{center}
\includegraphics[width=1\textwidth]{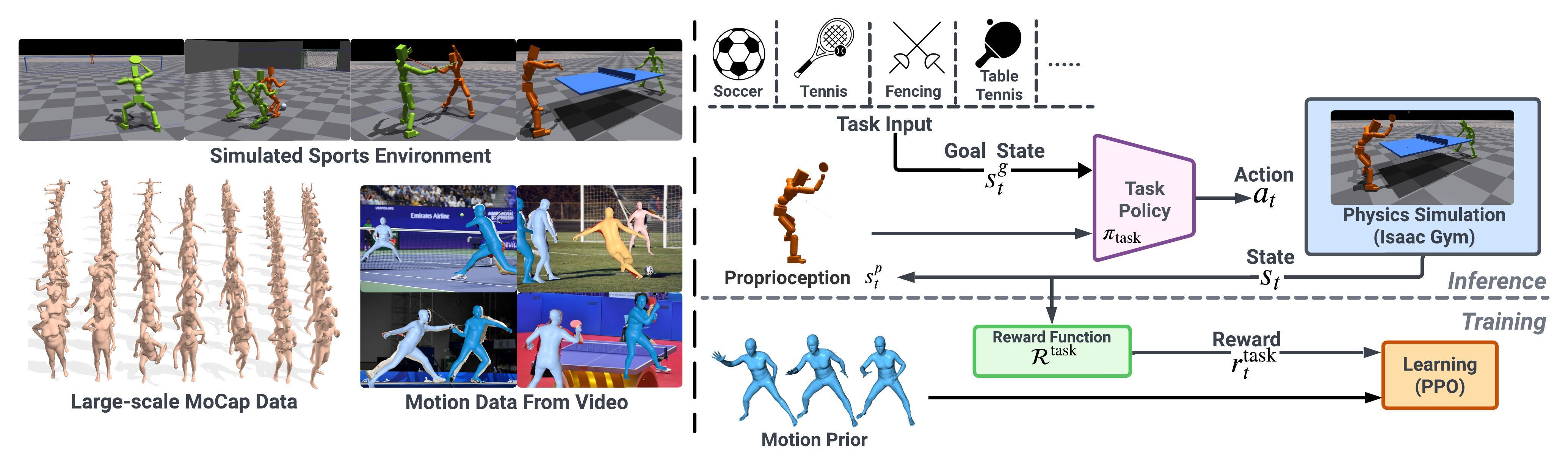}
\end{center}
\vspace{-5mm}
   \caption{\small{An overview of $\name$: we design a collection of simulated sports environments and leverage RL and human demonstrations (from videos or MoCap) as prior to tackle them. }}
\vspace{-6.5mm}
\label{fig:archi}
\end{figure*}

In this section, we describe the formulation of each of our sports environments, from single-person sports (Sec.~\ref{sec:single}) to multi-person sports (Sec.~\ref{sec:multi}). Then, we describe our pipeline for acquiring human demonstration data from videos (Sec.~\ref{sec:data}). An overview can be found in Fig.~\ref{fig:archi}. For each sport, we provide a preliminary reward design that serves as a baseline for future research. Due to space constraints, omitted details can be found in the supplement.

\subsection{Single-person Sports}
\label{sec:single}
\paragraph{High Jump}  In the high jump environment, the humanoid jumps over a horizontal bar placed at a certain height without touching it and aims to reach a goal point that is 2 meters behind the bar. The bar is positioned following the setup of the official Olympic game. The high jump goal state $\goalstatehighjump = (\bs{p}^{b}_t, \bs{p}^{g}_t)$ contains the positions of the bar $\bs{p}^{b}_t \in \mathbb{R}^{3}$ and the goal point $\bs{p}^{g}_t \in \mathbb{R}^{3}$. The reward is defined as $\rewardfunchighjump(\selfstate,\goalstatehighjump) \triangleq w^{\text{p}}r^{\text{p}}_t  + w^{\text{h}}r^{\text{h}}_t $. The position reward $r^{p}_t$ encourages the humanoid to go closer to the goal point. The height reward $r^{h}_t$ encourages the humanoid to jump higher. Training terminates when the humanoid is in contact with the bar, does not pass the bar, or falls to the ground before jumping. We also set up four bar heights for curriculum learning: 0.5m, 1m, 1.5m, and 2m.

\paragraph{Long Jump} Long jump is also set similar to the real-world setting, with a 20m runway followed by a jump area. Before the humanoid jumps, its feet should be behind the jump line. The goal state $\goalstatelongjump \triangleq (\bs{p}^{s}_t, \bs{p}^{l}_t, \bs{p}^{g}_t) $ includes the position of the starting point $\bs{p}^{s}_t \in \mathbb{R}^{3}$, jump line $\bs{p}^{l}_t \in \mathbb{R}^{3}$, and the goal $\bs{p}^{g}_t \in \mathbb{R}^{3}$. The training reward is defined as $\rewardfunclongjump(\selfstate, \goalstatelongjump) \triangleq w^{\text{p}}r^{\text{p}}_t + w^{\text{v}}r^{\text{v}}_t + w^{\text{h}}r^{\text{h}}_t  + w^{\text{l}}r^{\text{l}}_t$. The position reward $r^{\text{p}}_t$ encourages the humanoid to get closer to the goal, the velocity reward $r^{\text{v}}_t$ encourages larger running speed, and the height reward $r^{\text{h}}_t$ encourages higher jump. Finally, $r^{\text{l}}_t$  encourages jumping far.

\paragraph{Hurdling} In hurdling, the humanoid tries to reach a finishing line 110 meters ahead and needs to jump over 10 hurdles (each 1.067m high, placed 13.72m from the start, with subsequent hurdles spaced every 9.14m). The goal state is defined as $\goalstatehurdling \triangleq (\bs{p}^{h}_{t}, \bs{p}^{f}_{t})$, where $\bs{p}^{h}_{t} \in \mathbb{R}^{10\times3}$ and $\bs{p}^{f}_{t} \in \mathbb{R}^{3}$ includes the positions of these hurdles as well as the finish line. We define a simple reward function as $\rewardfunchurdling (\selfstate, \goalstatehurdling) \triangleq r^{\text{distance}}_t$. $\rewardfunchurdling$ encourages the agent to run towards the finish line and clear each hurdle. Additionally, we employ a curriculum for hurdling, where the height of each hurdle is randomly sampled between 0 and 1.167 meters for each episode.

\paragraph{Golf} For golf, the humanoid's right hand is replaced with a golf club measuring 1.14 meters. The driver of the golf club is simulated as a small box ($0.05\text{m} \times 0.025\text{m} \times 0.02\text{m}$). We incorporate a wave-like terrain with an amplitude of 0.5 meters in the golf environment, designed to mimic real-world grasslands. The golf goal is positioned to the left of the humanoid, at a distance ranging from 0 meters to 20 meters away. The goal state $\goalstategolf \triangleq ( \bs{p}^{b}_{t}, \bs{p}^{c}_{t}, \bs{p}^{g}_{t}, \bs{o}_t)$ includes the ball position $ \bs{p}^{b}_{t} \in \mathbb{R}^{3}$, club $ \bs{c}^{b}_{t} \in \mathbb{R}^{3}$, goal position $ \bs{p}^{g}_{t} \in \mathbb{R}^{3}$, and terrain height map $\bs{o}_t \in \mathbb{R}^{32 \times 32}$. The reward is defined as $\rewardfuncgolf (\selfstate, \goalstategolf) \triangleq w^{\text{p}}r^{\text{p}}_t +w^{\text{c}}r^{\text{c}}_t + w^{\text{g}}r^{\text{g}}_t +w^{\text{pred}}r^{\text{pred}}_t$, where the $r^{\text{p}}_t$ encourages the ball to move forward, $r^{\text{c}}_t$ encourages swinging the golf club to hit the ball, and $r^{\text{g}}_t$ encourages the ball to reach the goal. In addition, we predict the ball's trajectory and provide a dense reward $r^{\text{pred}}_t$ based on the distance between the predicted landing point and the goal.

\paragraph{Javelin} For javelin throw, we use SMPL-X humanoid with articulated fingers. 
The goal state is defined as $\goalstatejavelin \triangleq (\objp, \bs{p}^{r}_t, \bs{p}^{h}_t)$, where $\objp \in \mathbb{R}^{13}$, includes the position, orientation, linear, and angular velocity of the javelin. $\bs{p}^{r}_t$ and $\bs{p}^{h}_t$ are the positions of the root and right hand. The reward is defined as $\rewardfuncjavelin (\selfstate, \goalstatejavelin) \triangleq w^{\text{grab}}r^{\text{grab}}_t+w^{\text{js}}r^{\text{js}}_t+w^{\text{goal}}r^{\text{goal}}_t+w^{\text{s}}r^{\text{s}}_t$. The grab reward $r^{\text{grab}}_t$ encourages the right hand to grab the javelin. The javelin stability reward $r^{\text{js}}_t$ minimizes the javelin's self-rotation. The goal reward $r^{\text{goal}}_t$ encourages the humanoid to throw the javelin further. The stability reward $r^{\text{s}}_t$ is to avoid large movements of the body.

\subsection{Multi-person Sports}
\label{sec:multi}
\paragraph{Tennis} For tennis, each humanoid's right hand is replaced as an oval racket.  We use the same measurement as a real tennis court and ball. We design two tasks: a single-player task where the humanoid trains to hit balls launched randomly, and a 1v1 mode where the humanoid plays against another humanoid. For both tasks, the goal state is defined as $\goalstatetennis \triangleq (\ballp, \ballv, \racketp, \targetp$, where $\ballp \in \mathbb{R}^{3}, \ballv \in \mathbb{R}^{3}, \racketp \in \mathbb{R}^{3}$ and $\targetp \in \mathbb{R}^{3}$, which includes the position and velocity of the ball, position of the racket and position of the target. The reward function for tennis is defined as $\rewardfunctennis (\selfstate, \goalstatetennis) \triangleq w_{\text{p}}\rewardracket + w_{\text{b}}\rewardball$. The racket reward $\rewardracket$ encourages the racket to reach the ball, and the ball reward $\rewardball$ aims to successfully hit the ball into the opponent's court, as close to the target as possible. For the single-player task, we shoot a ball from the opposite side from a random position and trajectory, simulating a ball hit by the opponent. The target $\targetp$ is also randomly sampled. For the 1v1 scenario, we can either train models from scratch or initialize two identical single-player models as opponents, which can play back and forth. 

\paragraph{Table Tennis} For table tennis, each humanoid is equipped with a circular paddle (replacing the right hand) and plays on a standard table. Similar to tennis, we have the single-player task and the 1v1 task. Similarly, the goal state is defined as $\goalstatetennis \triangleq (\ballp, \ballv, \racketp, \targetp)$. The reward function for table tennis is defined as $\rewardfunctabletennis (\selfstate, \goalstatetabletennis) \triangleq w_{\text{p}}\rewardracket + w_{\text{b}}\rewardball$. The paddle reward $\rewardracket$ is the same as tennis while we modify the $\rewardball$ slightly to encourage more hits for table tennis.

\paragraph{Fencing} For 1v1 fencing, each humanoid is equipped with a sword (replacing the right hand) and plays on a standard fencing field.  The goal state is defined as $\goalstatefencing \triangleq (\oppt, \oppv, \simtsword - \oppttarget, \| \contact \|_2^2, \| \contactopp \|_2^2, \bbox )$, which contains the opponent's position body $\oppt \in \mathbb{R}^{24 \times 3}$, linear velocity $\oppv \in \mathbb{R}^{24 \times 3}$, the difference between target body position $\oppttarget \in \mathbb{R}^{5 \times3}$ on the opponent and agent's sword tip position $\simtsword$, normalized contract forces on the agent itself $\| \contact \|_2^2 \in \mathbb{R}^{ 24 \times 3}$ and its opponent $\| \contactopp \|_2^2 \in \mathbb{R}^{24 \times 3}$, as well as the bounding box $\bbox \in \mathbb{R}^{4} $. To train the fencing agent, we define the fencing reward function as $\rewardfuncfencing (\selfstate, \goalstatefencing) \triangleq w_{\text{f}}\rewardfacing + w_{\text{v}}\rewardvel + w_{\text{s}}\rewardstrike + w_{\text{p}}\rewardpoint$. The facing $r^{\text{facing}}_t$ and velocity reward $r^{\text{vel}}_t$ encourage the agent to face and move toward the opponent. The strike  reward $r^{\text{strike}}_t$ encourages the agent's sword tip to get close to the target, while $r^{\text{point}}_t$ is the reward for getting in contact with the target. We use the pelvis, head, spine, chest, and torso as the target bodies. The episode terminates if either of the humanoids falls or steps out of bounds. 

\paragraph{Boxing} For boxing, we simulate two humanoids with sphere hands in a bounded arena. The goal state is similar to fencing: $\goalstateboxing \triangleq (\oppt, \oppv, \simthand - \oppttarget, \| \contact \|_2^2, \| \contactopp \|_2^2)$ but without the bounding box information. The reward function and target body parts are also the same as fencing, though replacing the sword tip to the hands. 

\paragraph{Soccer} The soccer environment includes one or more humanoids, a ball, two goal posts, and the field boundaries. The field measures 32m $\times$ 20m. We support three tasks: penalty kicks, 1v1, and 2v2.

For penalty kicks, the humanoid is positioned 13 meters from the goal line, with the ball placed at a fixed spot 12 meters directly in front of the goal center. The objective is to kick the ball toward a randomly sampled target within the goal post. To achieve this, the controller is provided  $\goalstatekick \triangleq (\simtball, \simballv, \simtgoalpost, \simtgoaltarget)$, where $\simtball \in \mathbb{R}^{3}$ is the ball position, $\simballv \in \mathbb{R}^{3}$ is the velocity and angular velocity, $\simtgoalpost \in \mathbb{R}^{4}$ is the bounding box of the goal, and $\simtgoaltarget \in \mathbb{R}^{3}$ is the target location within the goal post. The reward is $\rewardfuncsoccerpenalty (\selfstate, \goalstatekick) \triangleq w^\text{p2b}r^\text{p2b} + w^\text{b2g}r^\text{b2g} + w^\text{bv2g}r^\text{bv2g} + w^\text{b2t}r^\text{b2t} - c^\text{no-dribble}_t$. Various rewards are designed to guide the character towards a run-and-kick motion. The player-to-ball ($r^\text{p2b}$) reward motivates the character to move towards the ball. The ball-to-goal reward ($r^\text{b2g}$) reduces the distance between the ball and the target. The ball-velocity-to-goal ($r^\text{bv2g}$) encourages a higher velocity of the ball toward the target. The ball-to-target ($r^\text{b2t}$) reward encourages a smaller distance between the target and the predicted landing spot of the ball based on its current position and velocity. Finally, a negative reward ($c^\text{no-dribble}_t$) is applied if the character passes the spawn position of the ball, which discourages dribbling and encourages kicking.

Beyond penalty kicks, we explore team-play dynamics, including 1v1 and 2v2. The controller is provided with a state defined as $\goalstatesoccer \triangleq (\simtball, \simballv, \simtgoalpost, \allytroot, \opptroot)$, where $\allytroot \in \mathbb{R}^{3}$ and $\opptroot \in \mathbb{R}^{3}$ are the root positions of the ally and opponents (1 or 2). The controller is then trained using the following reward $\rewardfuncsoccertwovtwo (\selfstate, \goalstatesoccer) \triangleq w^\text{p2b}r^\text{p2b} + w^\text{b2g}r^\text{b2g} + w^\text{bv2g}r^\text{bv2g} + w^\text{point}r^\text{point}$, where $r^\text{p2b}$, $r^\text{b2g}$ and $r^\text{bv2g}$ are the same as in penalty kick. $r^\text{b2g}$ and $r^\text{bv2g}$ are zeroed out when the distance to the ball is greater than 0.5m. $r^\text{point}$, the scoring a goal, provides a one-time bonus and or penalty for goals. Notice that this is a rudimentary reward design compared to prior art \cite{Liu2021-iz} and serves as a starting point for further development.

\paragraph{Basketball} Our basketball environment is set up similarly to the soccer environment except for using the SMPL-X humanoid. The court measures 29m $\times$ 15m, with a 3m high hoop. We also introduce the task of free-throw, where the humanoid begins at a distance of 4.5 meters from the hoop with the ball initially positioned close to its hands. The objective is to successfully throw the basketball into the hoop. The goal state for this task is defined similarly to that of the soccer penalty kicks, with the distinction being the prohibition of foot-to-ball contact to maintain basketball rules.

\paragraph{Competitive Self-play} In competitive sports environments, we implement a basic adversarial self-play mechanism where two policies, initialized randomly, compete against each other to optimize their rewards. We adopt an alternating optimization strategy from \cite{Won2021-sn}, where one policy is frozen while the other is trained. This encourages each policy to develop offensive and defensive strategies, contributing to more competitive behavior, as observed in boxing and fencing ({\tt\small \href{\link}{supplement site}}).

\subsection{Acquiring Human Demonstration From Videos}
\label{sec:data}
We utilize TRAM~\cite{wang2024tram} for 3D motion reconstruction from Internet videos, providing robust global trajectory and pose estimation under dynamic camera movements, commonly found in sports broadcasting. Specifically, TRAM estimates SMPL parameters~\cite{Loper2015-ey} which include global root translation, orientation, body poses, and shape parameters. We further apply PHC~\cite{Luo2023-ft}, a physics-based motion tracker, to imitate these estimated motions, ensuring physical plausibility. We find these corrected motions are significantly more effective as positive samples for adversarial learning compared to raw estimated results. More details and ablation are provided in the supplementary materials.

\begin{figure*}
\vspace{-4mm}
\begin{center}
\includegraphics[width=1\textwidth]{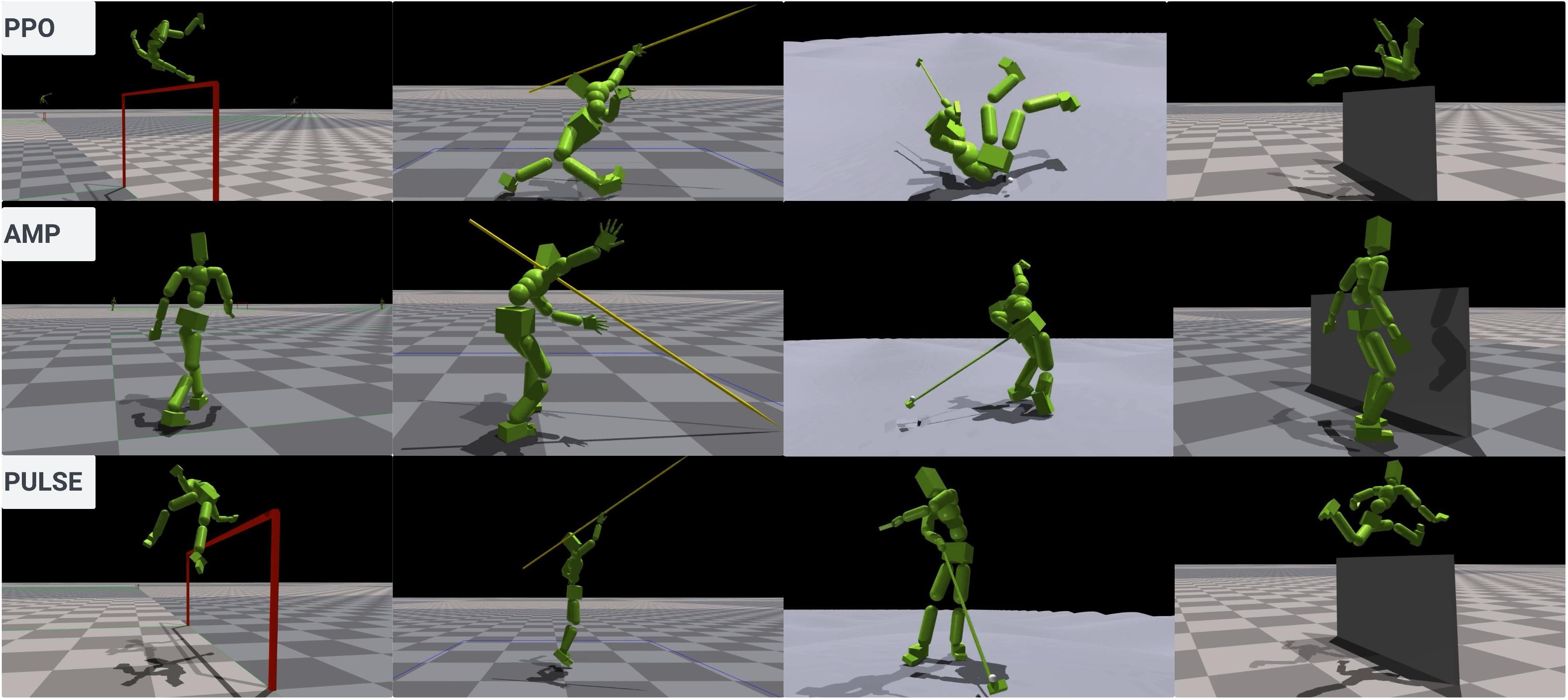}
\end{center}
\vspace{-4mm}
   \caption{\small{Qualitative results for high jump, javelin, golf, and hurdling. PPO and AMP try to solve the task using inhuman behavior, while PULSE can discover human-like behavior.}}
\vspace{-3.mm}
\label{fig:qual}
\end{figure*}

\section{Experiments}
\label{sec:exp}
\paragraph{Implementation Details} Simulation is conducted in Isaac Gym \citep{Makoviychuk2021-sx}, where the policy runs at 30 Hz and the simulation at 60 Hz. All task policies utilize three-layer MLPs with units [2048, 1024, 512]. The SMPL humanoid models adhere to the SMPL kinematic structure, featuring 24 joints, 23 of which are actuated, yielding an action space of $\mathcal{R}^{69}$. The SMPL-X humanoid has 52 joints, 51 actuated, including 21 body joints and hands, resulting in an action space of $\mathcal{R}^{153}$. Body parts on our humanoid consist of primitives such as capsules and blocks. All models can be trained on a single Nvidia RTX 3090 GPU in 1-3 days. We limit all joint actuation forces to 500 Nm.  For more implementation details, please refer to the supplement. 

\paragraph{Baselines} We benchmark our simulated sports using some of the state-of-the-art simulated humanoid control methods. While not a comprehensive list, it provides a baseline for the challenging environments. Each task is trained using PPO \cite{Schulman2017-ft}, AMP \cite{Peng2021-xu}, PULSE \cite{Luo2023-er}, and a combination of PULSE and AMP. AMP use a discriminator with the policy to provide an adversarial reward, using human demonstration data to deliver a ``style" reward that reflects the human-likeness of humanoid motion. Both task and discriminator rewards are equally weighted at 0.5. PULSE extracts a 32-dimensional universal motion representation from AMASS data, surpassing previous methods \cite{Tessler_undated-zi, Peng2022-vr} in coverage of motor skills and applicability to downstream tasks. Compared to AMP, PULSE uses hierarchical RL and offers a learned action space that accelerates training and provides human-like motion prior (instead of a discriminative reward). PULSE and AMP can be combined effectively, where PULSE provides the action space and AMP provides task-specific style reward.

\paragraph{Metrics} We provide quantitative evaluations for tasks with easily measurable metrics such as high jump, long jump, hurdling, javelin, golf, single-player tennis, table tennis, penalty kicks, and free throws. These metrics are detailed in the supplementary materials, where we also present qualitative assessments for tasks that are more challenging to quantify, such as boxing, fencing, and team soccer. Specifically, success rate (Suc Rate) determines whether an agent completes a sport according to set rules. Average distance (Avg Dis) indicates the extent an agent or object travels. For sports involving ball hits, such as tennis and table tennis, we record the average number of successful ball strikes (Avg Hits). Error distance (Error Dis) measures the distance between the intended target and the actual landing spot, applicable in sports like golf, tennis, and penalty kicks. Additionally, the hit rate in golf quantifies the success of striking the ball with the club. Evaluations are performed on 1000 trials.

\subsection{Benchmarking Popular Simulated Humanoid Algorithms}
\label{sec:benchmarking}
\begin{figure*}
\vspace{-1.mm}
\begin{center}
\includegraphics[width=1\textwidth]{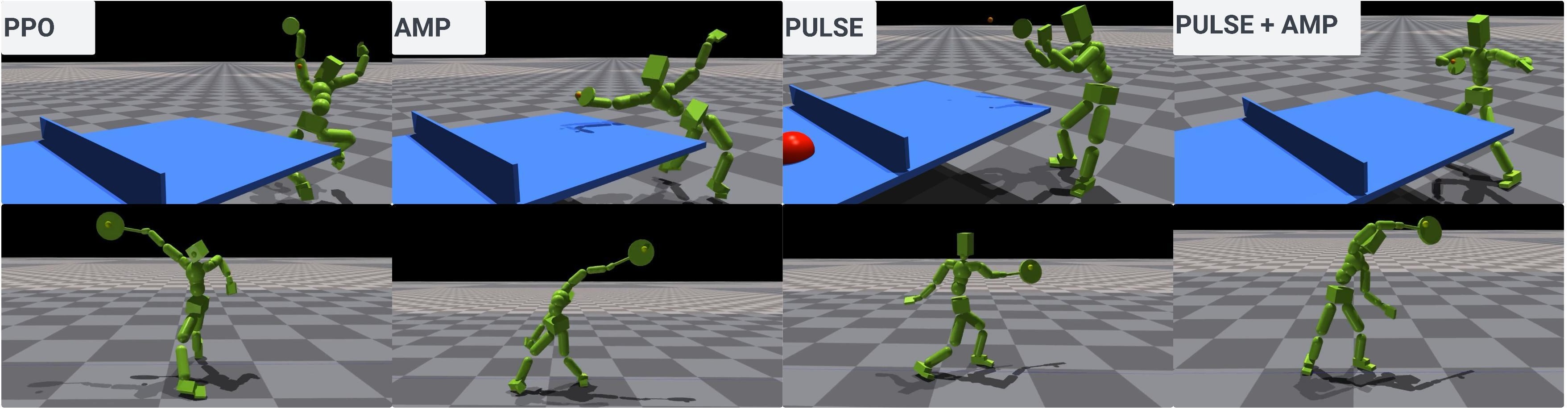}
\end{center}
\vspace{-4mm}
   \caption{\small{Qualitative results for table tennis and tennis. PPO and AMP result in inhuman behavior; PULSE can use human-like movement but PULSE + AMP result in behavior specific to the sport.}}
\vspace{-3.5mm}
\label{fig:qual2}
\end{figure*}

\begin{figure*}
\begin{center}
\includegraphics[width=1\textwidth]{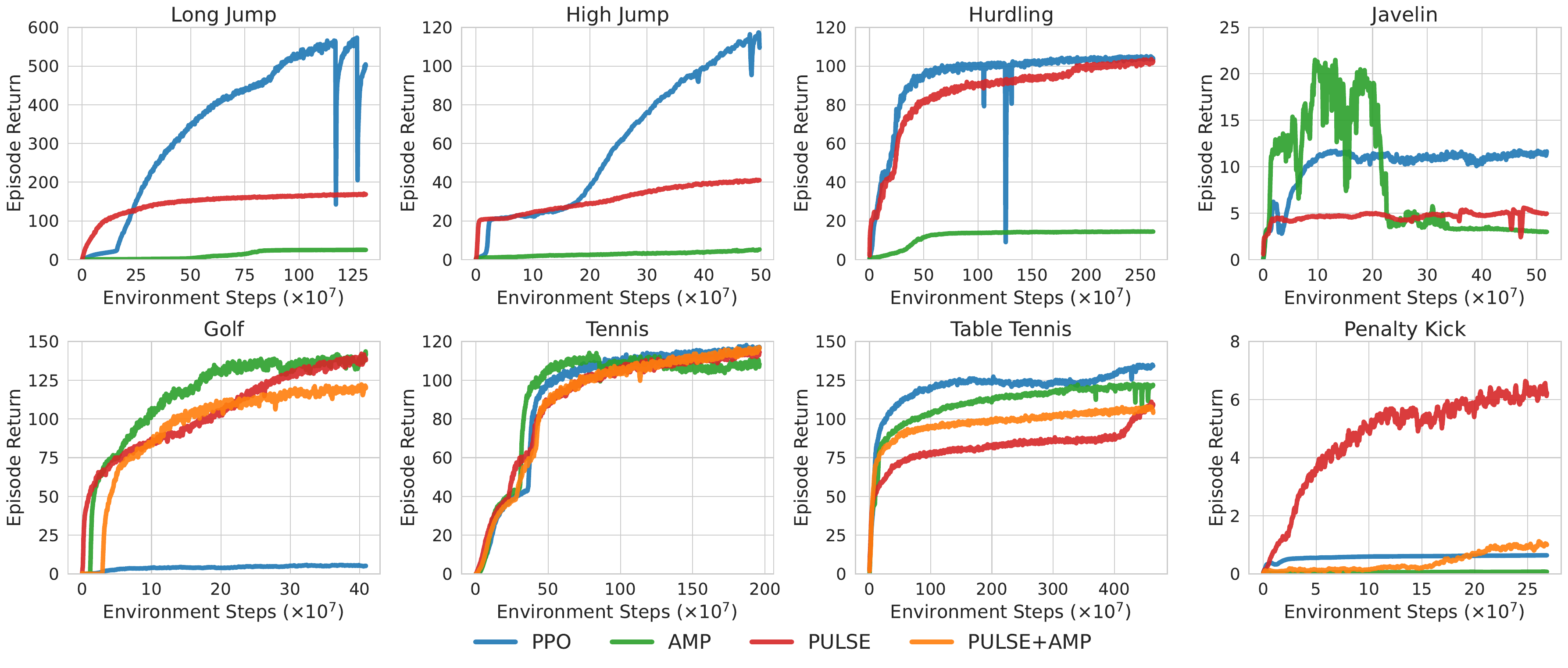}
\end{center}
\vspace{-0.2in}
   \caption{\small{Learning curves on various tasks. }}
\label{fig:curve}
\vspace{-0.2in}
\end{figure*}

In this section, we evaluate the performance of various control methods across our sports environments. We provide qualitative results in Fig.~\ref{fig:qual} and Fig.~\ref{fig:qual2}, and training curves in Fig.~\ref{fig:curve}. To view extensive qualitative results, including human-like soccer kick, boxing, high jump, \etc, please see {\tt\small \href{\link}{supplement}}. 

\paragraph{Track \& Field Sports (Without Video Data)}
We first evaluate track and field sports, including long jump, high jump, hurdling, and javelin throwing. For these sports, SOTA pose estimation methods fail to estimate coherent motion and global root trajectory as players and cameras are both fast-moving. Thus, we utilize a subset of the AMASS dataset containing locomotion data \cite{Rempe2023-tf} as reference motions. Since PULSE is pretrained on AMASS, we exclude PULSE + AMP from these tests. Table~\ref{tab:jumps_and_hurdle} summarizes the quantitative results of different methods. In long jump, AMP fails entirely, often walking slowly to the jump line without a forward leap. This failure occurs because the policy prioritizes discriminator rewards over task completion. If the task is too hard, the policy will use simple motion (such as standing still) to maximize the discriminator reward instead of trying to complete the task. In contrast, PPO, while capable of jumping great distances, exhibits unnatural motions. PULSE successfully executes jumps with human-like motion, but lacks the specialized skills for top-tier records due to the absence of corresponding motion data in AMASS. The high jump displays similar patterns: PPO achieves impressive heights but with unnatural movements while AMP struggles to reconcile adversarial and task rewards. Surprisingly, as shown in Figure~\ref{fig:qual}, PULSE successfully adopts a Fosbury flop approach without specific rewards to encourage this technique, likely leveraging breakdance skills. For hurdling, AMP completely fails, stopping before the first hurdle. PPO bounces energetically over each obstacle as shown in Figure~\ref{fig:qual}, but sometimes falls and fails to complete the race, with an average success rate of just over 50\% and an average distance of less than 110m. PULSE facilitates natural clearance of hurdles, and completes races in 17.76 seconds, a competitive time compared to human standards. Javelin throwing poses similar challenges: PPO uses inhuman strategies, AMP struggles with balancing rewards, and PULSE adopts human-like strategies but lacks specific skills for record-setting performance.

\begin{table*}[t]

\centering
\caption{\small{Evaluation on Long Jump, High Jump, Hurdling and Javelin. World records are in parentheses.}} 
\vspace{-0.05in}
\resizebox{\linewidth}{!}{%
\begin{tabular}{l|rr|rrrr|rrr|rr}
\toprule
\multicolumn{1}{c}{} & \multicolumn{2}{c}{Long Jump (8.95m)} & \multicolumn{4}{c}{High Jump (2.45m)} & \multicolumn{3}{c}{Hurdling (12.8s)}  & \multicolumn{2}{c}{Javelin (104.8m)} 
\\ 
\midrule
Method  & Suc Rate $\uparrow$ & Avg Dis $\uparrow$  & Suc Rate (1m) $\uparrow$ & Height (1m) $\uparrow$ & Suc Rate (1.5m) $\uparrow$ & Height (1.5m) $\uparrow$ & Suc Rate $\uparrow$& Avg Dis $\uparrow$ & Time $\downarrow$ & Suc Rate $\uparrow$ & Avg Dis $\uparrow$\\ \midrule

PPO~\cite{Schulman2017-ft} & 53.6\% & \textbf{19.42}   & \textbf{100\%} & \textbf{4.08} & \textbf{100\%} & \textbf{4.11}   & 57.6\% & 108.9 & \textbf{11.22} &  \textbf{100\%} & \textbf{44.5} \\
AMP~\cite{Peng2021-xu}     &  0\% & - & 0\% & - & 0\% & - & 0\% & 13.24 & - & 0.31\% & 2.03   \\
PULSE~\cite{Luo2023-er}    & \textbf{100\%} & 5.105   & \textbf{100\%} & 2.01 & \textbf{100\%}  & 1.98   & \textbf{100\%} & \textbf{122.1} & 17.76 & \textbf{100\%} & 9.63  \\
\bottomrule 
\end{tabular}}
\vspace{-0.15in}
\label{tab:jumps_and_hurdle}
\end{table*}

\begin{table*}[t]
\centering
\caption{\small{Evaluation on Golf, Tennis, Table Tennis, Penalty Kick and Free Throw}} 
\vspace{-0.05in}
\resizebox{\linewidth}{!}{%
\begin{tabular}{l|rr|rr|rr|rr|r}
\toprule
\multicolumn{1}{c}{}  &  \multicolumn{2}{c}{Tennis}  & \multicolumn{2}{c}{Table Tennis} & \multicolumn{2}{c}{Golf} & \multicolumn{2}{c}{Penalty Kick}  & \multicolumn{1}{c}{Free Throw}
\\ 
\midrule 
Method   & Avg Hits $\uparrow$ & Error Dis $\downarrow$ & Avg Hits $\uparrow$ & Error Dis $\downarrow$ & Hit Rate $\uparrow$ & Error Dis $\downarrow$  & Suc Rate $\uparrow$ & Error Dis $\downarrow$ & Suc Rate $\uparrow$ \\ \midrule

PPO~\cite{Schulman2017-ft}  & 2.76  & \textbf{1.92} & 1.01 &  \textbf{0.06} & 0\% & - & 0.0\% & - & \textbf{91.4\%}\\
AMP~\cite{Peng2021-xu}   & \textbf{3.95} &{ 5.30}  & 1.10 & {0.13} & \textbf{100\%} & 1.43 & 0.0\% & - & 0.0\%\\
PULSE~\cite{Luo2023-er}   & 2.48 & 3.50 & 0.74 & 0.19   & 99.9\% & \textbf{1.29} & \textbf{76.6\%} & \textbf{0.25} & {85.6\%} \\
PULSE~\cite{Luo2023-er} + AMP \cite{Peng2021-xu}  & 2.62 & 3.64 & \textbf{1.83} & 0.23 & 99.9\% & 2.18 & 27.5\% & 0.27 & {89.8}\% \\

\bottomrule 
\end{tabular}}
\vspace{-0.15in}
\label{tab:sports_with_video}

\end{table*}

\paragraph{Sports With Video Data}
For sports including golf, tennis, table tennis, and soccer penalty kick, we utilize processed motion from videos as demonstrations for AMP and PULSE+AMP. The results are reported in Table~\ref{tab:sports_with_video} and Fig.~\ref{fig:qual2}. In tennis, AMP demonstrates superior performance in terms of average hits; however, returned balls often land far from the intended targets. This is because prolonged rallies increase discriminator rewards, leading AMP to ignore task rewards. Notably, AMP exhibits inhuman motions at the moment of ball contact and reverts to natural movements when preparing for the next hit as shown in Fig.~\ref{fig:qual2}. This behavior underscores a reward conflict between balancing task and discriminator rewards. PPO plays tennis in an unnatural way, while PULSE and PULSE + AMP show similar performance. In table tennis, PPO achieves impressive error distances, but struggles with consistency and often fails to return second shots. We observe video data proves \emph{particularly beneficial for table tennis}. PULSE+AMP records significantly higher hit averages with reasonable error distances. Table tennis requires quick reactions within a short time, which the pre-trained PULSE model supports by providing necessary motor skills, enhanced by video data that guide the learning of proper stroke techniques. For golf, penalty kicks, and free throws, the ``initiating contact with an object" part makes them challenging. Here, only PULSE and PULSE+AMP manage to solve the three tasks effectively and \textit{consistently}, leveraging PULSE's latent space for effective exploration. The design of these tasks often results in a sparse exploration phase where triggering penalty rewards, such as $c^\text{no-dribble}_t$ for moving past the ball's initial position. The AMP reward also negatively affects training penalty kick, as the human demonstration contains other soccer motions such as running and dribbling, and the policy finds them easier to learn and exploit.


\paragraph{Curriculum learning} 
We find curriculum learning is an essential component in achieving better results for some tasks. In Table~\ref{tab:curriculum}, we study variants of high jump and hurdling task with and without
\begin{wrapfigure}{t}{0.5\linewidth}
\vspace{-3mm}
\centering
\makeatletter\def\@captype{table}\makeatother
\caption{\small{Evaluation on curriculum learning.}} 
\vspace{-2mm}
\resizebox{\linewidth}{!}{%
\begin{tabular}{l|rr|rrr}
\toprule
\multicolumn{1}{c}{} & \multicolumn{2}{c}{High Jump } & \multicolumn{3}{c}{Hurdling} 
\\ 
\midrule
Method  & Suc Rate (1m) & Suc Rate (1.5m) & Suc Rate & Avg Dis & Time \\ \midrule

w/o curriculum  & \textbf{100\%} & 0\% & 0\% & 13.65 & - \\
w/ curriculum & \textbf{100\%} & \textbf{100\%} & \textbf{100\%} & \textbf{122.1} & \textbf{17.76}  \\
\bottomrule 
\end{tabular}}
\vspace{-3mm}
\label{tab:curriculum}
\end{wrapfigure}
the curriculum using PULSE. We can see that without curriculum, high jump and hurdling both fail to solve the task. This is due to the policy not being able to obtain any reward facing challenging heights of bars and hurdles and the policy gets stuck in the local minima.

\section{Limitations, Conclusion and Future Work}
\label{sec:conclu}

\paragraph{Limitations } While $\name$ provides a large collection of simulated sports environments, it is far from being comprehensive. Certain sports are omitted due to simulation constraints (e.g., swimming, shooting, ice hockey, cycling) or their inherent complexity (e.g., 11-a-side soccer, equestrian events). Nevertheless, our framework is highly adaptable, allowing easy incorporation of additional sports like climbing, rugby, wrestling \etc. Our initial design of rewards, though able to achieve sensible results, is also far from optimal. For competitive sports such as 2v2 soccer and basketball, our results also fall short of SOTA \cite{Liu2021-iz} which employs much more complex systems. 

\paragraph{Conclusion and Future Work} We introduce $\name$, a collection of sports environments for simulated humanoids. We provide carefully designed state and reward, and benchmark humanoid control algorithms and motion priors. We find that by combining simple reward design and powerful human motion prior, one can achieve human-like behavior for solving various challenging sports. Our humanoid's compatibility with the SMPL family of models also provides an easy way to obtain additional data from video for training, which we demonstrate to be helpful in training some sports. These well-defined simulation environments could also serve as valuable platforms for frontier models \cite{ma2023eureka} to gain physical understanding. We believe that $\name$ provides a valuable starting point for the community to further explore physically simulated humanoids.

\bibliographystyle{plainnat} %
\bibliography{egbib}

\newpage

\appendix
{   
    \hypersetup{linkcolor=black}
    \begin{Large}
        \textbf{Appendix}
    \end{Large}
    \etocdepthtag.toc{mtappendix}
    \etocsettagdepth{mtchapter}{none}
    \etocsettagdepth{mtappendix}{subsection}
    \newlength\tocrulewidth
    \setlength{\tocrulewidth}{1.5pt}
    \parindent=0em
    \etocsettocstyle{\vskip0.5\baselineskip}{}
    \tableofcontents
}

\section{Introduction}

In the appendix, we provide comprehensive implementation details for $\name$, including the reward designs for each sport environment, training procedures, and hyperparameters. Extensive qualitative results can be accessed on our {\tt\small \href{\link}{supplement site}}, where we provide visualizations of all sports environments and training results based on our preliminary reward designs. Baseline results (PPO, AMP, PULSE, PULSE+AMP) are presented to support the quantitative findings discussed in the main paper. Furthermore, we offer visualizations of the reference motion extracted from in-the-wild videos. For our pipeline to acquire the human demonstration in the SMPL format, we conduct an ablation study evaluating the impact of employing a motion imitator (PHC \cite{Luo2023-ft}) as a refinement step. Code, videos, and asset attributions can also be found in our supplementary materials.

\section{Implementation Details}

\subsection{Rewards and Termination Conditions}

\paragraph{High Jump} For high jump, the humanoid's task is to leap over a horizontal bar positioned 20m ahead and 6m to the left of its starting point. The humanoid aims to reach the goal point $\highjumpposgoal= (22, 6, 1)$ located 2 m behind the bar. The reward function is defined as follows: 
\begin{equation}
    \rewardfunchighjump(\selfstate,\goalstatehighjump) \triangleq
        \begin{cases} 
                    1 \times r^{\text{p}}_t                              & \text{if \ } \highjumpposx \leq 19.5\text{m},  \\ 
                     1 \times r^{\text{p}}_t + 1 \times r^{\text{h}}_t   & \text{if \ } 19.5\text{m} < \highjumpposx < 20.5\text{m}, \\ 
                     1 \times r^{\text{p}}_t                              & \text{if \ } 20.5\text{m} \leq \highjumpposx.    \\ 
                  
        \end{cases}
\end{equation}
where $\highjumpposx$ denotes the x-axis position. The height reward, $r^{\text{h}}_t = \highjumpposz$, with $\highjumpposz$ representing the z-axis position, incentivizes the humanoid to jump higher. The position reward, $r^{\text{p}}_t  = \|\highjumpposprev - \highjumpposgoal\|_2 - \|\highjumppos - \highjumpposgoal\|_2$ (clamped to [0,1]), motivates the humanoid to reach the goal. An episode is terminated if the humanoid falls down, fails to leap over the bar, or moves beyond the designated run-up area. 

\paragraph{Long Jump} In the long jump environment, the humanoid has a 20-meter runway before the jump line, which its feet should not exceed.  The humanoid's goal is to reach the goal position, $\longjumpposgoal = (30, 0, 1)$. The training reward is defined as follows:  
\begin{equation}
    \rewardfunclongjump(\selfstate, \goalstatelongjump) \triangleq
        \begin{cases} 
                    1 \times r^{\text{p}}_t + 0.01 \times r^{\text{v}}_t             & \text{if \ } \longjumpposx \leq 20 \text{m}, \\ 
                    1 \times r^{\text{p}}_t + 0.01 \times r^{\text{v}}_t   + 0.1 \times r^{\text{h}} + 30 \times r^{\text{l}}           & \text{if \ } 20 \text{m} < \longjumpposx. \\ 
        \end{cases}
\end{equation}
The position reward, $r^{\text{p}}_t  = \|\longjumpposprev - \longjumpposgoal\|_2 - \|\longjumppos - \longjumpposgoal\|_2 $  (clamped to [0,1]) encourages the humanoid to reach the goal point. The velocity reward, $r^{\text{v}}_t = \longjumpvelocityx$ prompts the humanoid to reach higher speed along the x-axis. The jump height reward $r^{\text{h}}_t = \longjumpposz $ encourages the humanoid to jump higher after reaching the jump line. 
The jump length reward $r^{\text{l}}_t = \longjumpposx - 20 $ promotes longer final jump length. Each episode terminates if the humanoid falls or runs off the track.

\paragraph{Hurdling}  In the hurdling task, the humanoid aims to reach a finish line 110m ahead while jumping over 10 hurdles, each 1.067m high. The first hurdle is placed 13.72m from the start, with subsequent hurdles spaced every 9.14m. The reward function is defined as $\rewardfunchurdling (\selfstate, \goalstatehurdling) \triangleq r^{\text{distance}}_t$, which encourages the agent to run towards the finish line and clear each hurdle.  
\begin{equation}
    \rewardfunchurdling (\selfstate, \goalstatehurdling)  \triangleq   1 \times r^{\text{distance}}_t  \\ 
\end{equation}
The distance reward, $r^{\text{distance}}_t  = \|\hurdlingposprev - \hurdlingposgoal\|_2 - \|\hurdlingpos - \hurdlingposgoal\|_2$, is clamped to $[0,1]$ and encourages the humanoid to get closer to the goal point. We terminate each episode if the character falls or runs off the track.

\paragraph{Golf} In the golf task, the humanoid is equipped with a golf club of dimensions of $0.05\text{m} \times 0.025\text{m} \times 0.02\text{m}$. The target location for the golf ball is positioned to the left of the humanoid, in the direction of the x-axis, at a distance ranging from 0m to 20m. The reward function is defined as follows:   
\begin{equation}
    \rewardfuncgolf (\selfstate, \goalstategolf) \triangleq 1 \times r^{\text{p}}_t + 1 \times r^{\text{c}}_t + 1 \times r^{\text{g}}_t +1 \times r^{\text{pred}}_t \\    
\end{equation}
The position reward, $r^{\text{p}}_t  \triangleq  \|\ballpprev - \targetp\|_2 - \| \ballp - \targetp\|_2 $, clamped such that $0<r^{\text{p}}_t <1$, encourages the ball to get closer to the target. The contact reward $r^{\text{c}}_t$ encourages swinging the golf club to hit the ball, defined as:
\begin{equation}
    r^{\text{c}}_t =
 \begin{cases}
     1 \times  \exp(-100 \times \|\ballp - \clubp \|^2) & \text{if \ } C_{\text{cb}}=0, \\
      1 & \text{if \ } C_{\text{cb}}=1.
 \end{cases}
\end{equation}
Here, $C_{\text{cb}}=0$ indicates that the club has not made contact with the ball and $C_{\text{cb}}=1$ indicates the club has made contact. The goal reward, $ r^{\text{g}}_t = \exp (-0.1 \times \| \ballpxy - \targetpxy\|^2)$, encourages the ball to reach the target position in the x-y plane. In addition, we predict the ball's trajectory and provide a dense reward $r^{\text{pred}}_t = \exp (-0.1 \times \|\bs{p}^{\text{land}} - \ballpxy \|^2) $ based on the distance between the predicted landing point and the goal on the x-y plane~\cite{Zhang2023-ox}. The landing position, ${\bs{p}^{\text{land}}} = \left( x^{\text{land}}, y^{\text{land}} \right)$, can be calculated using the initial position and velocity as follows ($g$ is gravity):
\begin{equation}
    x_{\text{land}} = x_0 + v_{0,x} \left( \frac{v_{0,z} + \sqrt{v_{0,z}^2 + 2gz_0}}{g} \right) ,\ \ y_{\text{land}} = y_0 + v_{0,y} \left( \frac{v_{0,z} + \sqrt{v_{0,z}^2 + 2gz_0}}{g} \right) 
\end{equation}
 Early termination is triggered if the ball moves backward, does not contact the golf club within 2 seconds, is too close to the humanoid's body, or the humanoid falls. 
 
\paragraph{Javelin} For javelin throw, the humanoid is equipped with a javelin of length 2.7m. Due to the complexity introduced by articulated fingers, the reward function $\rewardfuncjavelin$ is applied in three stages: first, the humanoid learns to hold the javelin stably; then, it learns to throw it; finally, the javelin flies as far as possible. A timer is used to differentiate the three stages. Specifically,  $\rewardfuncjavelin$ is defined as follows:
\begin{equation}
    \rewardfuncjavelin (\selfstate, \goalstatejavelin) \triangleq 
        \begin{cases} 
                    0.9 \times r^{\text{grab}}_t + 0.1 \times r^{\text{js}}_t & \text{if \ } t < 0.6s,\\ 
                     0.9 \times r^{\text{goal}}_t + 0.05 \times r^{\text{s}}_t - 0.05 \times r^{\text{grab}}_t & \text{if \ }
 0.6s \leq t < 1.2s, \\ 
                     0.9 \times r^{\text{goal}}_t + 0.1 \times r^{\text{js}}_t & \text{if \ } 1.2s \leq t. \ 
        \end{cases}
\end{equation}
The reward for grasping $r^{\text{grab}}_t = \exp(-1 \times \| \righthandpos -  \javelinpos \|^2)$ encourages the hand to stay close to the javelin. The javelin stability reward $r^{\text{js}}_t = \exp(-1 \times \|\javelinpose - \javelinposedefault \| ^2)$ encourages the 6 DoF pose of the javelin to remain close to the default pose, which faces forward and tilts 30 degrees upward, mimicking a flying pose. The humanoid stability reward, $r^{\text{s}}_t = \exp(-1 \times \| \rootpos \|^2) $, encourages the humanoid to keep its root position fixed. The termination conditions vary according to the stage: during the grasping and throwing stages, the episode terminates if the javelin is too far from the right hand or deviates significantly from the default pose $\javelinposedefault$. During the flying stage, termination occurs if the javelin is too close to the right hand. 
 
\subsection{Multi-person Sports}
\label{sec:multi}
\paragraph{Tennis} For tennis, each humanoid is equipped with a circular racket with a 15cm radius, positioned 35cm away from the wrist, replacing the right hand. The court measures 23.77m in length and 8.23m in width, mirroring the dimensions and layout of a real tennis court. The net height is 1m, and the simulated ball has a radius of 3.2cm. We design two tasks: a single-player ball return task, where the humanoid trains to hit balls launched randomly, and a 1v1 mode, where the humanoid competes against another humanoid. In the ball return task, the humanoid is positioned at the center of the baseline, with balls launched from the opposite side. The landing location is uniformly sampled on the opposite side and the ball launch velocity is randomly sampled. The reward function is defined as follows:
\begin{equation}
\rewardfunctennis (\selfstate, \goalstatetennis) \triangleq  
   \begin{cases} 
    1 \times \rewardracket + 0 \times \rewardball,  & \text{if \ } C_{\text{rb}}=0, \\ 
    0 \times \rewardracket + 1 \times \rewardball, & \text{if \ } C_{\text{rb}}=1.
    \end{cases}
\end{equation}
Here, $C_{\text{rb}}=0$ indicates that the racket has not made contact with the ball, and $C_{\text{rb}}=1$ indicates the racket has made contact. $\rewardracket = \exp(-1 \times \|\racketp - \ballp \|^{2})$ rewards the racket for getting closer to the ball. $\rewardball = 1 + \exp(-1 \times \|\bs{p}^{\text{land}} - \targetp \|^{2})$ encourages the predicted landing location of the ball to be close to the target. Similar to the golf task, the landing location of the ball is calculated based on $\ballp$ and $\ballv$, providing a dense reward function to facilitate training~\cite{Zhang2023-ox}. Early termination occurs if the humanoid loses the point, either by failing to catch the ball or by hitting the ball out of bounds. In the 1v1 mode, two humanoids are placed on opposite sides of the court and the first ball is launched from the middle of the court, randomly directed at each player. The same reward function as the ball return task is used. To facilitate 1v1 training, the pre-trained model from the ball return task is used as a warm start. Similarly, the episode terminates if one player fails to catch the ball or returns the ball out of bounds.

 \paragraph{Table Tennis} For table tennis, each humanoid is equipped with a circular paddle with an 8 cm radius, positioned 12 cm from the wrist, replacing the right hand. The table adheres to standard dimensions, featuring a playing surface 2.74 m in length and 1.525 m in width, standing 0.76 m high. The net is 15.25 cm high, and the table tennis ball has a radius of 2 cm. The setup includes a single-player ball return task and a 1v1 task. The reward function is designed similarly to tennis, except we define the ball reward as $\rewardball = 1 + \exp(-1 \times \| \bs{p}^{\text{land}}- \targetp \|^{2}) + N_{\text{hit}}$, where $N_{\text{hit}}$ counts the number of successful hits in one episode. This formulation is intended to encourage the humanoid to continuously hit the ball effectively. Unlike in golf and tennis, we calculate $\bs{p}^{\text{land}}$ when it lands on the table at a height of 0.76 m. For early termination and the warm start in 1v1, we maintain the same setting as in the tennis task.

\paragraph{Fencing} For 1v1 fencing,  similar to real-world fencing, the two players are confined to a 14m by 2m playground, where stepping out of the bound will reset the game. The fencing reward is structured similarly to the boxing setup in NCP \cite{Zhu2023-nn}:
\begin{equation}
    \rewardfuncfencing (\selfstate, \goalstatefencing) \triangleq 0.1 \times \rewardfacing + 0.1 \times 
 \rewardvel + 0.6 \times  \rewardstrike + 1 \times \rewardpoint.
\end{equation} 

The facing reward $\rewardfacing $ penalizes deviation from facing the opponent's root position $\opptroot$. The velocity reward, $\rewardvel$, encourages the x-y plane linear velocity to be directed towards the opponent's root position $\opptroot$. The strike reward, $\rewardstrike = \exp(-10 \times \argmin \|\simtsword - \opptbodytarget \|^2)$,  encourages the swordtip to get closer to the target body parts $\opptbodytarget$, which include the pelvis, head, spine, chest, and torso. If there is contact with the target body part with sufficient force, a positive reward is provided:

\begin{equation}
    \rewardpoint = 
        \begin{cases} 
                    1  & \text{if \ } \argmin \|\simtsword - \opptbodytarget \|^2 \leq 0.1 \enskip \text{and} \enskip  \text{contact force} \geq 50 \text{Nm},\\ 
                     0 & \text{otherwise}.\\ 
        \end{cases}
\end{equation}
Our fencing agents are trained using competitive self-play, as introduced in the main paper.

\paragraph{Boxing} For boxing, the humanoid competes in a boxing ring measuring 5m by 5m. The humanoid's right hand is replaced with a sphere of 8cm radius. The boxing reward function has the same composition as fencing, except that the sword tip position $\simtsword$ is replaced by the hand position $\handpos$. Our boxing agents are also trained using competitive self-play.

\paragraph{Soccer} The soccer field measures 32m in length and 20m in width. Each goal is 4m wide and 2m tall.  The ball has a diameter of 11.5 cm and weighs 450 grams. For the penalty kick task, the reward function $\rewardfuncsoccerpenalty (\selfstate, \goalstatekick) \triangleq w^\text{p2b}r^\text{p2b} + w^\text{b2g}r^\text{b2g} + w^\text{bv2g}r^\text{bv2g} + w^\text{b2t}r^\text{b2t} - c^\text{no-dribble}_t$ is divided into stages based on whether the ball is moving toward the goal. Specifically, we define a "closer to goal" variable as $\closertogoalxy = \|\simtgoaltarget - \ballpprev \|_2  - \| \simtgoaltarget - \ballp \|_2$, which indicates whether the ball is getting closer to the goal. The full reward function is defined as follows:
\begin{equation}
    \rewardfuncsoccerpenalty (\selfstate, \goalstatekick) \triangleq 
    \begin{cases} 
         0.4 \times r^\text{p2b}  - c^\text{no-dribble}_t  & \text{if \ } \closertogoalxy \leq 0, \\
         0.1 \times r^\text{b2g} + 0.1 \times r^\text{bv2g} + 0.8 \times r^\text{b2t} - c^\text{no-dribble}_t & \text{otherwise}.\\
    \end{cases}
\end{equation}
Essentially, if the ball is not moving toward the goal, the humanoid is encouraged to move toward the ball; if the ball is moving, the agent is rewarded for shooting the ball toward the target in the goal post. The player-to-ball reward, $r^\text{p2b} =  \|\rootposprev - \ballpprev \|_2  - \| \rootpos - \ballp \|_2  $, is a point-goal reward \cite{Won2022-jy}. The ball-to-goal reward $r^\text{b2g} = \|\simtgoaltarget - \ballpprev \|_2  - \| \simtgoaltarget - \ballp \|_2$ encourages the ball to move closer to the goal position. The ball-velocity-to-goal reward $r^\text{bv2g}$ incentivizes the ball velocity toward the goal position. The ball-to-target reward $r^\text{b2t}$ predicts the landing position of the ball in the net based on its current velocity and position, providing a reward if the ball is close to the target. Finally, $c^\text{no-dribble}_t$ penalizes the humanoid if its root position is over the ball's spawning point. 

The team play (1v1 and 2v2) soccer tasks use similar rewards as the penalty kick task. The reward function for team play is $\rewardfuncsoccertwovtwo (\selfstate, \goalstatesoccer) \triangleq w^\text{p2b}r^\text{p2b} + w^\text{b2g}r^\text{b2g} + w^\text{bv2g}r^\text{bv2g} + w^\text{point}r^\text{point}$, where $r^\text{p2b}$, $r^\text{b2g}$ are the same as in the penalty kick. $r^\text{point}$ provides a one-time bonus for scoring.

\paragraph{Basketball} The basketball environment is similar to soccer except that it utilizes the SMPL-X humanoid with articulated fingers. In the free-throwing task, the ball is initialized between the humanoid's hands. The free throw reward is defined as: $\rewardfuncfreethrow (\selfstate, \goalstatesoccer) \triangleq 0.5 \times r^\text{ballvel} + 0.5 \times r^\text{bv2g} + r^\text{basket}$. The basketball velocity reward $r^\text{ballvel} =\exp(-0.1 \times  \|\simballlv - \simballlvdesired  \|_2^2)$ encourages the ball's velocity to be close to the desired velocity to reach the goal. The desired velocity, $\simballlvdesired$, is computed using the goal position $\simtgoaltarget$, and the ball position $\ballp$, with the following physics equations:
\begin{equation}
    \begin{aligned} 
    T^{\text{reach}}_t &= \sqrt{\frac{2 \times \|(\ballp - \simtgoaltarget)_z \|_2}{g}} \ , \
    \simballlvdesiredxy = \frac{\|(\ballp - \simtgoaltarget)_{xy} \|_2}{T^{\text{reach}}_t} 
    \\
    \simballlvdesiredz &= \frac{(\ballp - \simtgoaltarget)_z  + 0.5 \times g  \times (T^{\text{reach}}_t)^2}{T^{\text{reach}}}.
    \end{aligned}
\end{equation}
The ball-velocity-to-goal reward $r^\text{bv2g} $ encourages the velocity to be directed towards the goal position. The basket reward, $r^\text{basket}$, provides a one-time reward if the ball passes through the basket.

Team-play basketball has a similar reward design as soccer. The team-play basketball task is highly challenging due to the difficulty of picking the ball up, which is more complex than kicking a ball. Thus, while we support 1v1 and 2v2 team-play basketball, our preliminary reward design does not yield interesting behavior, unlike in soccer.

\subsection{Hyperparamters}
Training hyperparameters are provided in Table~\ref{tab:supp_hyper}. We use the same set of hyperparameters to train \textit{all} of our sports environments, highlighting the advantage of employing a unified humanoid embodiment for simulated sports. 

\begin{table}[t]
\centering
\caption{Hyperparameters for training each baseline used in $\name$. We use the same set of hyperparamters for \textit{each sport}. Notice that AMP and PULSE uses PPO as the optimization method but add respective motion priors (as reward or motion representation). $\sigma$: fixed variance for policy. $\gamma$: discount factor. $\epsilon$: clip range for PPO. $w_\text{disc}$ and $w_\text{task}$: weights for discriminator and task rewards. } 
\resizebox{\linewidth}{!}{%
\begin{tabular}{lccccccccc}
\toprule
    & Batch Size & Learning Rate   & $\sigma$ &  $\gamma$  & $\epsilon$  &  MLP-size & $w_\text{disc}$&$w_\text{task}$ &\# of samples
  \\ \midrule
    PPO \cite{Schulman2017-ft} &  1024 & $5 \times 10^{-4}$ &  0.05   &  0.99  & 0.2 & [2048, 1024, 512] & 0 & 1& $\sim 10^{9}$\\ 
    AMP~\cite{Peng2021-xu} &  1024 & $5 \times 10^{-4}$ &  0.05   &  0.99  & 0.2 & [2048, 1024, 512] & 0.5& 0.5&$\sim 10^{9}$\\ 
    PULSE~\cite{Luo2023-er} &  1024 & $5 \times 10^{-4}$ &  0.3   &  0.99  & 0.2 & [2048, 1024, 512] & 0 & 1&$\sim 10^{9}$\\ 
    PULSE~\cite{Luo2023-er} + AMP~\cite{Peng2021-xu} &  1024 & $5 \times 10^{-4}$ &  0.3   &  0.99  & 0.2 & [2048, 1024, 512] & 0.5 & 0.5 & $\sim 10^{9}$\\ 
\bottomrule 
\end{tabular}}\\ 
\label{tab:supp_hyper}
\end{table}

\subsection{Details about Baselines}
For our baseline methods PULSE \cite{Luo2023-er} and AMP \cite{Peng2021-xu}, we use the official implementations. For PULSE \cite{Luo2023-er}, we employ the publicly released model without modification, which is pre-trained on the AMASS dataset. We follow a similar setup for downstream tasks in PULSE, using the frozen prior $\prior$,  decoder $\decoder$, and residual action representation. Since PULSE only includes trained models for the SMPL-based models, we train SMPL-X humanoid based models following the official code. Specifically, we train a humanoid motion imitator following PHC \cite{Luo2023-ft}, and distill motor skills into a 48-dimensional latent space (instead of 32-D, to accommodate articulated fingers). PULSE provides an action space for hierarchical RL and can be integrated with AMP. For PULSE+AMP, the AMP reward offers additional style guidance for the humanoid, which is particularly beneficial for tasks such as table tennis. However, we find that the demonstration sequences used for AMP need to be task-specific (\eg contains only a swinging motion); otherwise, the discriminator reward can overpower the task reward and lead to undesired behavior (as seen in the free kick results).

\section{Additional Ablations}

We conducted an ablation study to evaluate the role of physics-based tracking (w/ PHC) in acquiring human reference motion. Specifically, we used the pose estimation results directly from TRAM~\cite{wang2024tram} as positive samples for the discriminator during policy training (w/ PHC). Our experiments were performed in the context of table tennis. As shown in Table~\ref{tab:phc}, we found that providing video data without PHC leads to significantly lower performance compared to using PHC, similar to the results obtained using only PULSE. We observe that when the quality of the provided reference motion is poor (e.g., with significant noise in position, 
\begin{wrapfigure}{t}{0.4\linewidth}
\vspace{-1mm}
\centering
\makeatletter\def\@captype{table}\makeatother
\caption{\small{Ablation study on PHC.}} 
\vspace{-2mm}
\resizebox{\linewidth}{!}{%
\begin{tabular}{l|rr}
\toprule
\multicolumn{1}{c}{} & \multicolumn{2}{c}{Table Tennis } 
\\ 
\midrule
Method  & Avg Hits $\uparrow$ & Error Dis $\downarrow$ \\ \midrule
PULSE & 0.74 & 0.19   \\
PULSE+AMP, w/o PHC  & 0.91 & \textbf{0.18}   \\
PULSE+AMP, w/ PHC & \textbf{1.83} & 0.23   \\
\bottomrule 
\end{tabular}}
\vspace{-6mm}
\label{tab:phc}
\end{wrapfigure}
and drastic velocity changes), the model struggles to effectively utilize the reference motion as style guidance to achieve natural movements. In contrast, employing physics-based tracking to refine pose estimates from in-the-wild videos results in physically plausible motion, which significantly aids in policy learning.

\section{Broader Social impact } We propose $\name$, a collection of sports environments for simulated humanoids. These environments can be used to benchmark learning algorithms, discover new humanoid behaviors, create animations, and more. The potential negative social impact includes the risk of generating animations that could be used to create DeepFakes. Positive social impact includes the development of intelligent and collaborative agents, advancements in robot learning, discovery of new sports techniques, and the generation of immersive and physically realistic animations.

\end{document}